\documentclass[letterpaper, 10 pt, conference]{ieeeconf}  

\IEEEoverridecommandlockouts                              

\usepackage[utf8]{inputenc} 
\usepackage[T1]{fontenc}    
\usepackage{hyperref}       
\usepackage{url}            
\usepackage{booktabs}       
\usepackage{amsfonts}       
\usepackage{nicefrac}       
\usepackage{microtype}      
\usepackage{xcolor}         
\usepackage{boldline}
\usepackage{arydshln}
\usepackage{multirow}
\usepackage{amsmath}
\usepackage{xspace}
\usepackage{caption}
\usepackage{longtable}
\usepackage{graphicx}
\usepackage{wrapfig}

\def\eg{\emph{e.g.}\xspace} 
\def\ie{\emph{i.e.}\xspace}

\newcommand{\Kai}[1]{{#1}}

\title{\LARGE \bf
\textit{EfficientEQA}: An Efficient Approach to Open-Vocabulary \\Embodied Question Answering
}
\author{{Kai Cheng$^{*1}$, Zhengyuan Li$^{*1}$ \thanks{*Equal contribution.}, Xingpeng Sun$^1$, Byung-Cheol Min$^1$, Amrit Singh Bedi$^2$, Aniket Bera$^1$}\\
{\textit{$^1$Purdue University, USA  $^2$University of Central Florida, USA}}\\
}

\begin{document}

\maketitle
\thispagestyle{empty}
\pagestyle{empty}

\begin{abstract}
Embodied Question Answering (EQA) is an essential yet challenging task for robot assistants. Large vision-language models (VLMs) have shown promise for EQA, but existing approaches either treat it as static video question answering without active exploration or restrict answers to a closed set of choices. These limitations hinder real-world applicability, where a robot must explore efficiently and provide accurate answers in open-vocabulary settings.
To overcome these challenges, we introduce EfficientEQA, a novel framework that couples efficient exploration with free-form answer generation. EfficientEQA features three key innovations: 
(1) Semantic-Value-Weighted Frontier Exploration (SFE) with Verbalized Confidence (VC) from a black-box VLM to prioritize semantically important areas to \textit{explore}, enabling the agent to gather relevant information faster; (2) a BLIP relevancy-based mechanism to \textit{stop} adaptively by flagging highly relevant observations as outliers to indicate whether the agent has collected enough information; and (3) {a} Retrieval-Augmented Generation (RAG) method for the VLM to \textit{answer} accurately based on pertinent images from the agent’s observation history without relying on predefined choices.
Our experimental results show that EfficientEQA achieves over 15\% higher answer accuracy and requires over 20\% fewer exploration steps than state-of-the-art methods.
Our code is available at: \href{https://github.com/chengkaiAcademyCity/EfficientEQA}{https://github.com/chengkaiAcademyCity/EfficientEQA}

\end{abstract}


\section{Introduction}

When entering an unknown scene, humans naturally explore the environment and can answer open-ended questions with ease. This capability has been formalized as Embodied Question Answering (EQA)~\cite{das2018embodied,gordon2018iqa, yu2019multi, das2018neural, wijmans2019embodied}. For robots, however, achieving such open-vocabulary EQA capability is extremely challenging due to the complexity of real-world scenes and the unbounded answer space. Despite recent advances in large vision-language models (VLMs) that offer powerful visual reasoning, most existing approaches either treat the problem as passive video question answering without active exploration \cite{majumdar2024openeqa,luo2022depth,cangea2019videonavqa} or constrain the answers to a closed set of options in simplified EQA settings ~\cite{exploreeqa2024,das2018embodied,gordon2018iqa}. These limitations severely restrict a robot’s ability to handle realistic, open-ended queries.

For example, consider a household robot tasked with answering the question \textit{$Q$: “How many cushions are there on the red sofa in the living room?”} (cf. Fig. \ref{fig:teaser}). To respond accurately, the robot must first locate the living room, identify the red sofa, and recognize what constitutes a cushion — all without prior knowledge of the room’s layout. This requires exploring its environment, deciding its current location, planning efficient navigation, and engaging in visual reasoning to extract relevant information.
We define this challenging task of open vocabulary embodied question answering during exploration as \textbf{Open Explore EQA} (cf. Fig. \ref{fig:teaser}). Addressing open-ended embodied QA with black-box VLMs introduces several challenges that extend beyond traditional EQA formulations:
 
\begin{figure}[tbp]
    \centering
    \includegraphics[width=\linewidth]{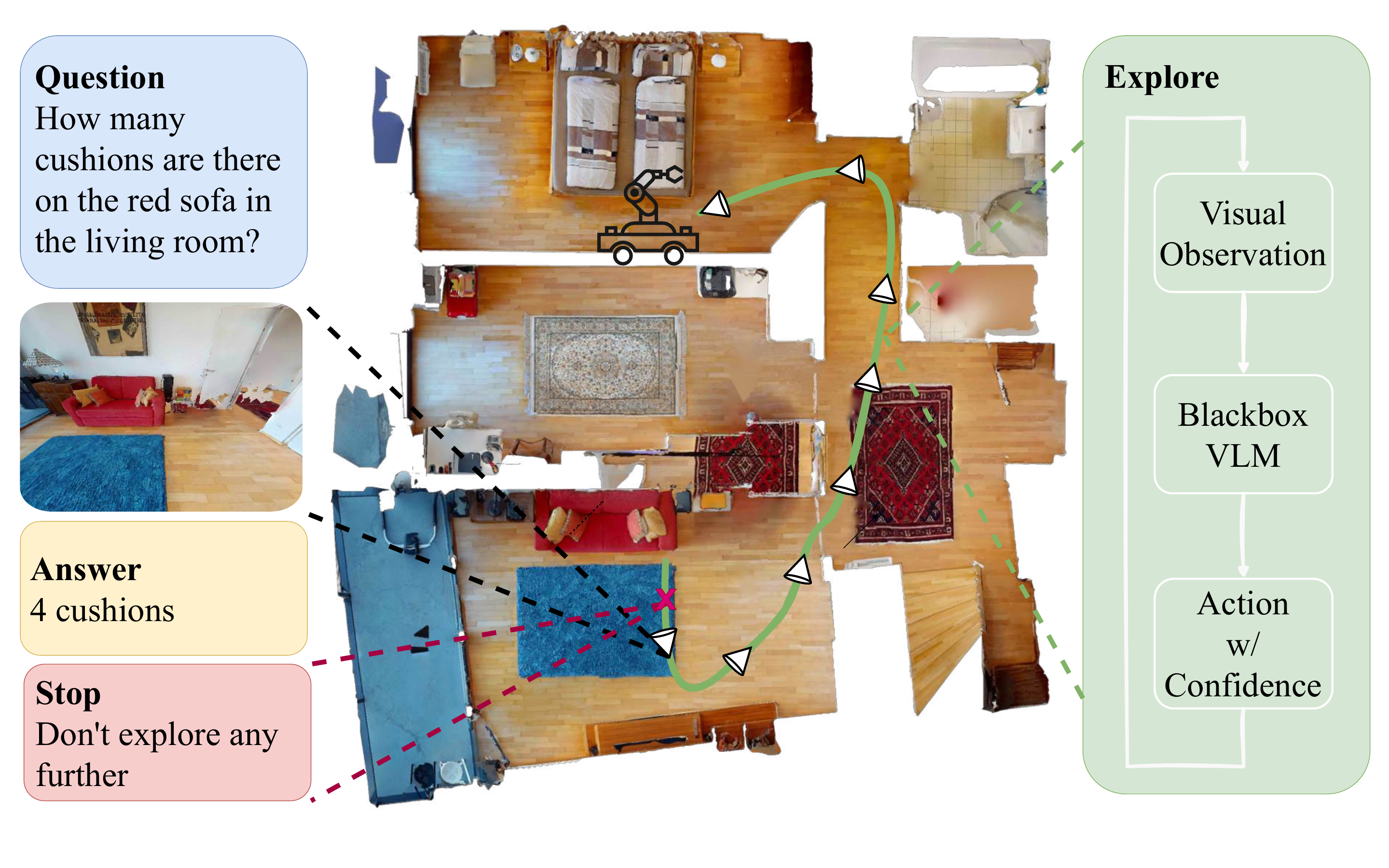} 
    \caption{\textbf{Our Proposed {Task Setting.} 
    } Given an open vocabulary question \textit{$Q$: "How many cushions are there on the red sofa in the living room?"} about the scene $\mathcal{S}$ which is unknown to the robot agent, our proposed model can actively \textbf{explore} the environment to get visual observations, which is then fed into a black-box VLM to predict the next-step action together with its confidence. Without any prior knowledge about the answer sets, our 
    {proposed model \textbf{EffiicientEQA}} will explore for efficient steps to \textbf{stop} when it is confident enough and then give an \textbf{answer} \textit{$A$: "4 cushions"}.}
    \label{fig:teaser}
    \vspace{-3mm}
\end{figure}

\begin{enumerate}
    \item[C1:] \textbf{Open-vocabulary answering.} How to search through the open vocabulary answer space without choice sets? 
    \item[C2:]  \textbf{Efficiently exploring unknown environments.} How to extract  {exploration confidence scores}
    of {exploration} directions 
    from VLMs (with black-box access)?
    \item[C3:] \textbf{Deciding when to stop searching.} How to decide when to stop exploring and whether the answer is accurate enough?
\end{enumerate}

\noindent To address these challenges, we introduce \textbf{EfficientEQA}, a novel framework that enables a robot agent to efficiently explore and accurately answer open-vocabulary embodied questions. Specifically, we divide a typical embodied question-answering process into three stages: to \textit{Explore}, \textit{Answer}, and \textit{Stop}, as shown in Fig.~\ref{fig:teaser}. In the stage to \textit{explore}, we use \textit{Semantic-Value-Weighted Frontier Exploration} (SFE)~\cite{yamauchi1997frontier} to build the semantic map and \textit{Verbalized Confidence} (VC) to extract 
exploration confidence scores from black-box VLMs. In the stage to \textit{answer}, we use Retrieval-Augmented Generation (RAG) to generate the predicted answer based on the observation episode history. For the stage to \textit{stop}, we use outlier detection to select the \textit{answer-decisive} frame as the stop sign for the entire exploration process. 
With the stopping module preventing unnecessary exploration and SFE prioritizing more informative regions, our exploration process focuses on areas most relevant to the given question. As a result, EfficientEQA effectively reduces redundant exploration steps, thus making it more \textit{efficient} in open-vocabulary EQA scenarios.
We summarize our contributions as follows:

\begin{itemize}
\item We introduce and formulate the task of \textbf{Open Explore EQA} (open vocabulary embodied question answering during exploration) with motivation from real-world scenarios.
\item We present a novel framework \textbf{EfficientEQA} to answer embodied questions based on Semantic-Value-Weighted Frontier Exploration (SFE) \cite{yamauchi1997frontier} and extracting exploration confidence scores from black-box VLMs.
\item We propose a RAG-based VLM prompting method and an outlier-detection-based early-stopping strategy to improve accuracy and efficiency.
\item We conduct extensive experiments using the Habitat-Sim physics simulator~\cite{puig2023habitat3} with the large-scale Habitat-Matterport 3D Research Dataset (HM3D)~\cite{ramakrishnan2021habitat} and HM-EQA~\cite{exploreeqa2024}, as well as real-world demonstrations in home/office-like environments using Husky and UR5e mobile manipulators. Experimental results show that EfficientEQA achieves over 15\% higher answer accuracy and requires over 20\% fewer exploration steps than state-of-the-art methods.
\end{itemize}

\section{Related Work}
 
\noindent \textbf{Embodied question answering.} EQA~\cite{das2018embodied,gordon2018iqa, yu2019multi, das2018neural, wijmans2019embodied} is a task in which an agent navigates an unfamiliar 3D environment to answer specific queries. {Unlike Visual Question Answering (VQA)~\cite{vqa,wu2017visual,kafle2017visual,coco-qa}, where the model directly infers answers from static images, EQA requires active exploration in the environment to gather the necessary visual information before answering.} Early works like~\cite{gordon2018iqa, yu2019multi} approached this by constructing holistic models entirely from scratch, leading to highly resource-intensive training and a time-consuming data collection process. More recent studies demonstrate that VLMs, such as GPT-4V~\cite{achiam2023gpt}, LLAVA~\cite{liu2023improvedllava}, and Prismatic VLM~\cite{karamcheti2024prismatic}, are capable of handling EQA tasks. Open-vocabulary EQA has been explored in~\cite{majumdar2024openeqa}, but their approach focuses more on video-based question answering instead of exploration. In contrast, \cite{exploreeqa2024} employed VLMs to both explore environments and answer questions. Our work diverges from theirs by leveraging advanced black-box {VLMs}, {with a focus}
on the more challenging and realistic open-vocabulary setting.

\vspace{2mm}
\noindent \textbf{{VLMs for embodied agents}.} There have been tremendous research endeavors studying VLMs' capacity for visual question answering and reasoning, which have shown great prospects for application to embodied tasks~\cite{pmlr-v229-huang23b, li2024manipllm, zhang2024navid,huang2023diffvl,ding2024open6dor,ma2022sqa3d, hong20233dmv, hong20233dllm, huang2023embodied}. The success of pre-trained open-vocabulary vision-language models~\cite{hong2024cogvlm2,radford2021learning,minderer2022simple} significantly boost the zero-shot performance of data-driven embodied agents in manipulation~\cite{li2024manipllm,liang2023code}, navigation~\cite{shah2023navigation,sun2024trustnavgpt,zhou2024navgpt}, and question-answering~\cite{exploreeqa2024} tasks. Fine-tuned VLMs have also exhibited strong physical and spatial reasoning capability.~\cite{chen2024spatialvlm, gao2023physically, brohan2023rt} These results further eliminate the high cost of data collection and fine-tuning when transferring to new problem domains. However, current VLM embodied agent works do not consider how agents can efficiently explore the environment while performing open-vocabulary EQA tasks in a training-free manner. Our paper distinctively fills this gap 
{by enabling agents to explore an unknown scene without modifying it.
Previous works like NLMap~\cite{chen2023open}, VLMap~\cite{huang2023visual}, CoW~\cite{gadre2022cow}, VLFM~\cite{yokoyama2024vlfm}, and HM3D-OVON~\cite{yokoyama2024hm3d} have attempted to extract object information from spatial maps obtained from pretrained VLMs to do navigation tasks. However, these navigation methods basically build spatial maps offline, and target locations can be queried simply given the text input. In our task setting, the agent has no prior knowledge of the scene and builds the semantic map online during exploration.}

\vspace{2mm}
\noindent \textbf{{Verbalized} confidence {from LLMs}.}
\begin{figure*}[h]
    \centering
    \includegraphics[width=\textwidth]{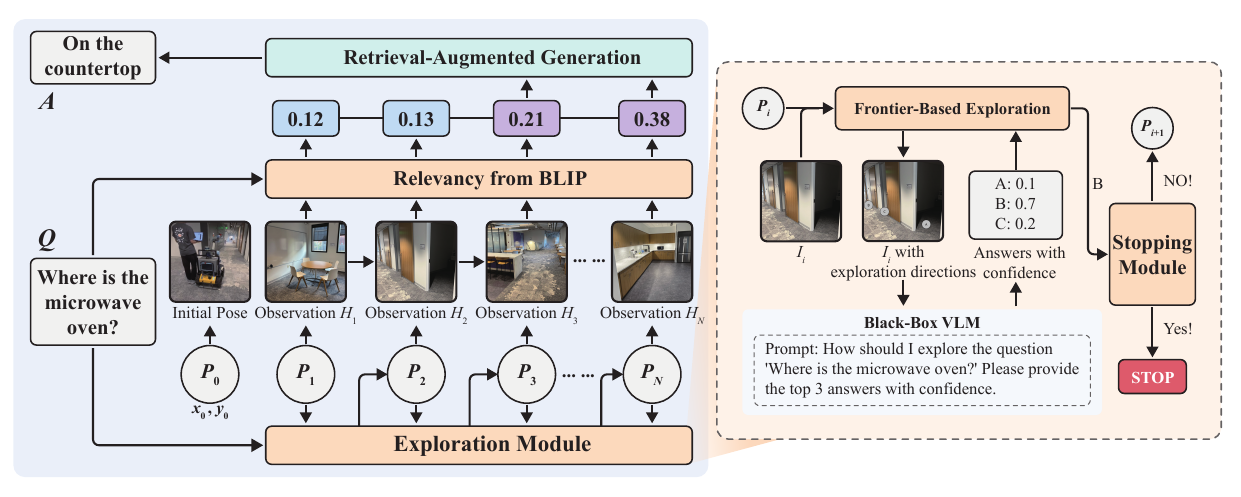} 
    \caption{\textbf{Our Proposed \textit{EfficientEQA} Framework.} Given the question \textit{$Q$: "Where is the microwave oven?"} as input, the robot agent explores $N$ steps to get observations. During exploration, the frontier-based exploration provides candidate directions ($A,B,C,...$) to the Black-Box VLM, which then evaluates each candidate's confidence \Kai{scores}. This evaluation determines the next step and updates the semantic map in frontier-based exploration. The agent doesn't stop until it observes an \textit{answer-decisive} frame $H^*$ when it knows the answer. The episode history $\{H_1, H_2, ...,H^*\}$ is filtered by BLIP and then fed into the VLM for retrieval-augmented generation to get the predicted answer \textit{$A^*$: "On the countertop"}. } 
    \label{fig:pipeline}
\end{figure*}
Measuring confidence in the outputs of language models is a fundamental challenge in machine learning.~\cite{kadavath2022language, mielke2022reducing,ren2023knowno} Early approaches rely on additional data and fine-tuning techniques to extract confidence from LLMs. For instance, \cite{mielke2022reducing} trains an auxiliary model to assess the certainty of language models, while \cite{lin2022teaching} fine-tunes GPT-3~\cite{brown2020language} to generate explicit verbalized confidence.
\Kai{Most state-of-the-art LLMs like~\cite{achiam2023gpt, team2023gemini} are usually accessed through the application user interface, where parameters are not available. While some open-source models~\cite{li2024llava, hong2024cogvlm2, karamcheti2024prismatic} offer comparable performance {with logit access}, their deployment in resource-constrained environments, such as household robots, is limited by the significant computational requirements they impose. To make advantage of such} 
closed-source LLMs, research has increasingly focused on estimating confidence solely from model outputs in a black-box setting. \cite{kuhn2023semantic} maps outputs to a semantic space and calculates entropy as an uncertainty estimate. \cite{su2024api} introduces a conformal prediction-based confidence measure for LLMs without logit access; however, this method provides confidence guarantees on prediction sets rather than assessing the correctness of individual responses, \Kai{which is required by our task}. Meanwhile, \cite{tian2023just} explores eliciting verbal confidence directly from LLMs, bypassing the need for sampling-based probability estimation. \cite{xiongcan} further formalizes a framework for prompting LLMs to elicit confidence, analyzing multiple strategies.
{Following~\cite{xiongcan}}, we 
{obtain the model's verbalized confidence by directly querying the}
vision-language model
{to guide our online exploration.}

\section{Problem Formulation}
We formulate a new challenging task, \textbf{Open Explore EQA}, for a mobile robot to explore a 3D scene $\mathcal{S}$ aiming to infer the answer $\mathit{A}$ of a given open-vocabulary embodied question $\mathit{Q}$. Each embodied question answering task $\mathit{\xi}$ is defined as a tuple $\xi = (\mathcal{S},\mathit{P}_0,\mathit{Q},\mathit{A}^*)$, where $\mathcal{S}$ is the given real or simulated 3D scene, $\mathit{P}_0$ is the initial pose of the robot, $\mathit{Q}$ is the open-vocabulary embodied question, and $\mathit{A}^*$ refers to the ground truth answer of $\mathit{Q}$. We assume no prior knowledge of the distribution of $\mathcal{S}$, $\mathit{Q}$, and $\mathit{A}^*$ so the robot agent must explore $\mathcal{S}$ from scratch to get the context of $\mathit{Q}$, \ie, the robot agent needs to take exploratory, information gathering actions to give the answer (\eg, \textit{‘Q: Do we have canned tomatoes at
home? A: Yes, I found canned tomatoes in the pantry.’}~\cite{majumdar2024openeqa}). {Note that our task does not require the robot agent to interact with surrounding objects, apart from exploration.}

{Starting from a random location $\mathit{P}_0$ in the scene $\mathcal{S}$, the robot needs to decide where to explore next based on visual input from a single camera installed in its front. Since there is no prior knowledge of the scene, the robot needs to build the environment map from scratch and actively explore around. In terms of exploration efficiency, the robot needs to decide which areas might be more related to the input question $\mathit{Q}$.}
While most previous EQA works~\cite{exploreeqa2024,wijmans2019embodied} mainly consider closed-form question answering, \ie, only answering questions $\mathit{Q}$ with a given finite answer choice set $\mathcal{Y} = \{\textrm`A\textrm', \textrm`B\textrm', \textrm`C\textrm', \textrm`D\textrm', ..., \textrm`Some\ Answer\textrm'\}$ where $\mathit{A}^* \in \mathcal{Y}$, we assume completely no knowledge of $\mathcal{Y}$ and perform question answering together with environment exploring in an open-vocabulary manner, which motivates the name Open Explore EQA.

\vspace{2mm}
\noindent \textbf{Remark.} We remark that the setting of Open Explore EQA is specifically challenging because there is no prior knowledge of $\mathcal{S}$, and $\mathit{A}^*$, which makes it difficult and time-consuming to explore the whole scene and search through the unknown answer distribution. In this work, we address this challenge and propose a method that efficiently explores the environment and stops as early as the agent is confident and answers the question accurately using its history of visual observations, as is shown in Fig.~\ref{fig:pipeline}. We next discuss our proposed method.

\section{Proposed Approach}

\noindent 
\textbf{Overview.} Our \textbf{EfficientEQA} framework is composed of three main modules: (A) the exploration module $\mathcal{E}$ (Sec.~\ref{sec:ex1},~\ref{sec:ex2}),  (B) the stopping module $\mathit{S}$ (Sec.~\ref{sec:st}), and (C) the answering module $\mathcal{A}$ (Sec.~\ref{sec:ans}). An overview of our pipeline is shown in Fig.~\ref{fig:pipeline}. Starting from the initial pose $P_0$, the robot agent actively explores the scene by capturing each step's observation $H_t = (I_t, D_t)$. For the exploring module $\mathcal{E}$, our model builds the environment map by fusing the observation history $H = \{H_1, H_2, \dots, H_t\}$ at step $t$ using Semantic-Value-Weighted Frontier Exploration (SFE)~\cite{exploreeqa2024,yamauchi1997frontier}. Then the exploring direction is predicted by Verbalized Confidence (VC) from black-box models. For the stopping module $\mathit{S}$, our model detects the \textit{answer-decisive} frame $H^*$ as an outlier from the distribution of the episode history. For the answering module $\mathcal{A}$, the episode history $\{H_1, H_2, ...,H^*\}$ is filtered by BLIP (Bootstrapping Language-Image Pre-training)~\cite{li2023blip} and then fed into the VLM for RAG to generate an answer.

\subsection{\textbf{Exploration - SFE:} Semantic-Value-Weighted Frontier Exploration}
\label{sec:ex1}
As the first component of our exploration module $\mathcal{E}$, Semantic-Value-Weighted Frontier Exploration (SFE)~\cite{exploreeqa2024,yamauchi1997frontier} provides an effective methodology for navigating unknown environments using semantic insights. This approach requires only a preliminary understanding of the scene's dimensions, and it dynamically updates an environmental map based on continual observations and corresponding actions.
Upon capturing a new observation $H_i$ at position $P_i$, the robot first identifies and selects voxels that are visible in the image according to the camera's positioning. These voxels are then marked as explored, and their semantic values are updated to reflect new data. Subsequently, the system proposes several potential paths to investigate the boundary between explored and unexplored areas. A VLM not only selects the most promising direction for further exploration, but also gauges confidence levels associated with each option. Thus, the semantic map is revised to incorporate the latest findings, serving as a guide for subsequent exploration. 

Finally, the robot advances to the designated location $P_{i+1}$, positioning itself for the next phase of exploration. Notably, the role of SFE is limited to map construction; it neither answers queries nor decides when to end the exploration process. SFE also requires the VLM to provide a semantic understanding of the observation, which we will elaborate upon in the next section.

\subsection{\textbf{Exploration - VC:} Verbalized Confidence from Black-Box Models} 
\label{sec:ex2}

As the second component of $\mathcal{E}$, the VLM is prompted to decide the next place to explore and provide a confidence score for each option. Because most state-of-the-art models are black boxes to users, we need to elicit verbalized confidence (VC) without knowing its internal states.
In order to 
{infer}
the confidence, we apply a pipeline that obtains $M$ answers and normalized confidence simultaneously, \Kai{ following~\cite{xiongcan}}.

\begin{figure}[t]
    \centering
    \includegraphics[width=\linewidth]{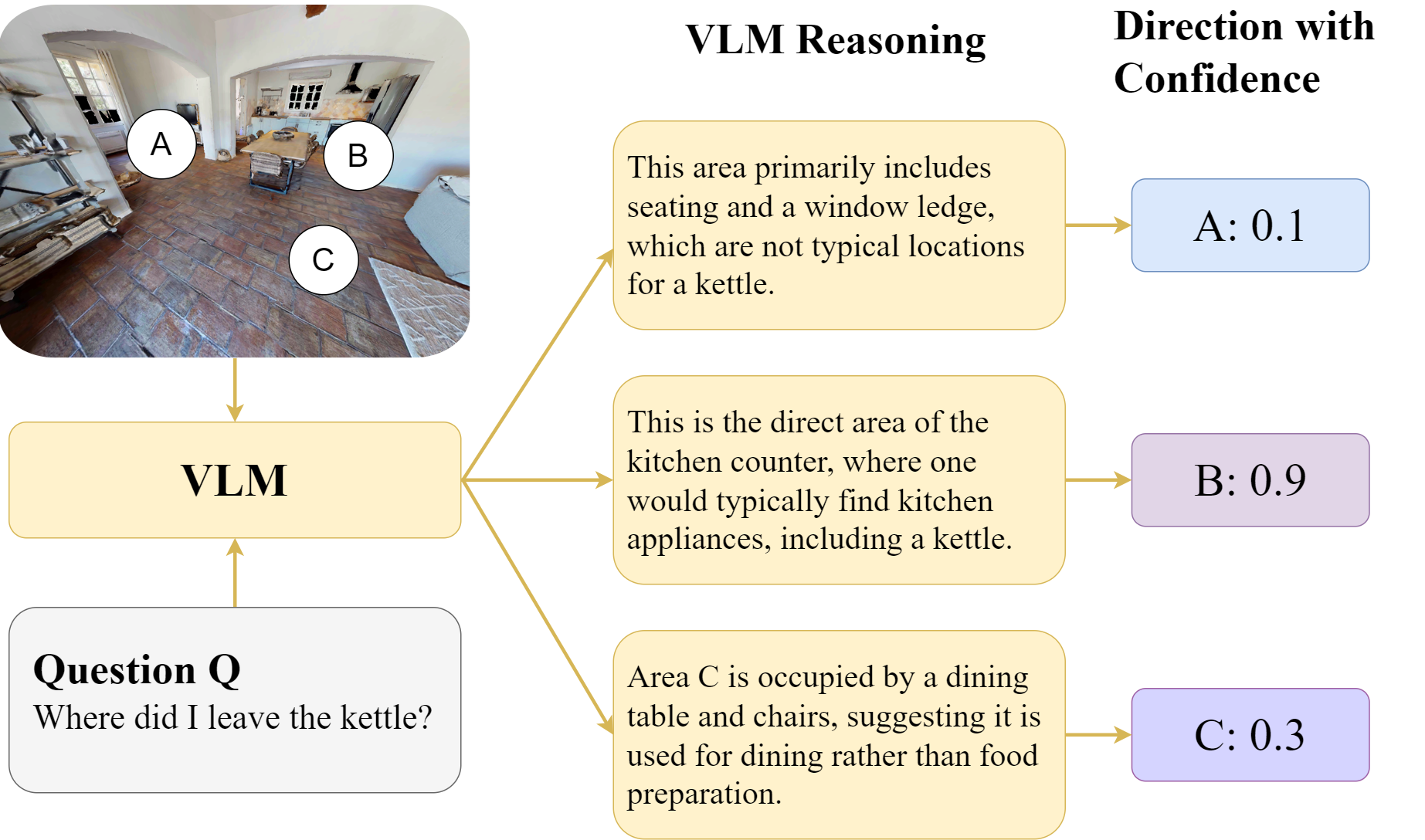} 
    \caption{\textbf{Verbalized Confidence for Exploration.} This visualization serves to show how the VLM reasons about space and provides guidance to exploration. In practice, the VLM directly outputs answers and associated confidence 
    \Kai{scores} without verbal reasoning. }
    \label{fig:confidence}
\end{figure}

First, we ask for $M$ answers to the question concurrently and then the resultant response is parsed, \Kai{as shown in Fig.~\ref{fig:confidence} (where $M=3$)}. We get a set of answers and corresponding confidence $\{r_i, c_i\}_{i=1}^{M}$. Note that the answers might overlap with others and the confidence is not normalized. Second, we normalize the confidence to ensure the sum is one:
$c_i' = c_i/\sum c_i.$
Third, we ask language models to annotate answers with the same meanings, we assume that all such answers are in the same sematic equivalent class $C$. By leveraging language models, we implicitly implement the semantic grouping function \( E(\cdot) \): 
\begin{equation}
    E(r_i) = E(r_j)\quad \text{iff.}\ r_i,r_j \in R_k.
\end{equation}
The above operation groups all answers into semantic equivalence classes \( \{R_k\} \) with cumulative confidences \( \{C_k\} \), where
\begin{equation}
    C_k = \sum_{R_k=E(r_i)} c_i'.
\end{equation}
Lastly, we select the semantic equivalence class with the highest confidence as
\begin{equation}
    (R^*,C^*) = \arg\max_{C_k} (R_k,C_k).
\end{equation}
We visualize the application of verbalized confidence to our method in Fig.~\ref{fig:confidence}. The exploration module $\mathcal{E}$ marks the locations to explore in observation $I_i$ and then the VLM is asked to provide confidence in each candidate's direction. Then those confidence scores are normalized and fed back to $\mathcal{E}$ to update the internal semantic map.

\subsection{\textbf{Stopping:} Relevancy-Based Early Stopping}
\label{sec:st}
\begin{figure}[t]
    \centering
    \includegraphics[width=\linewidth]{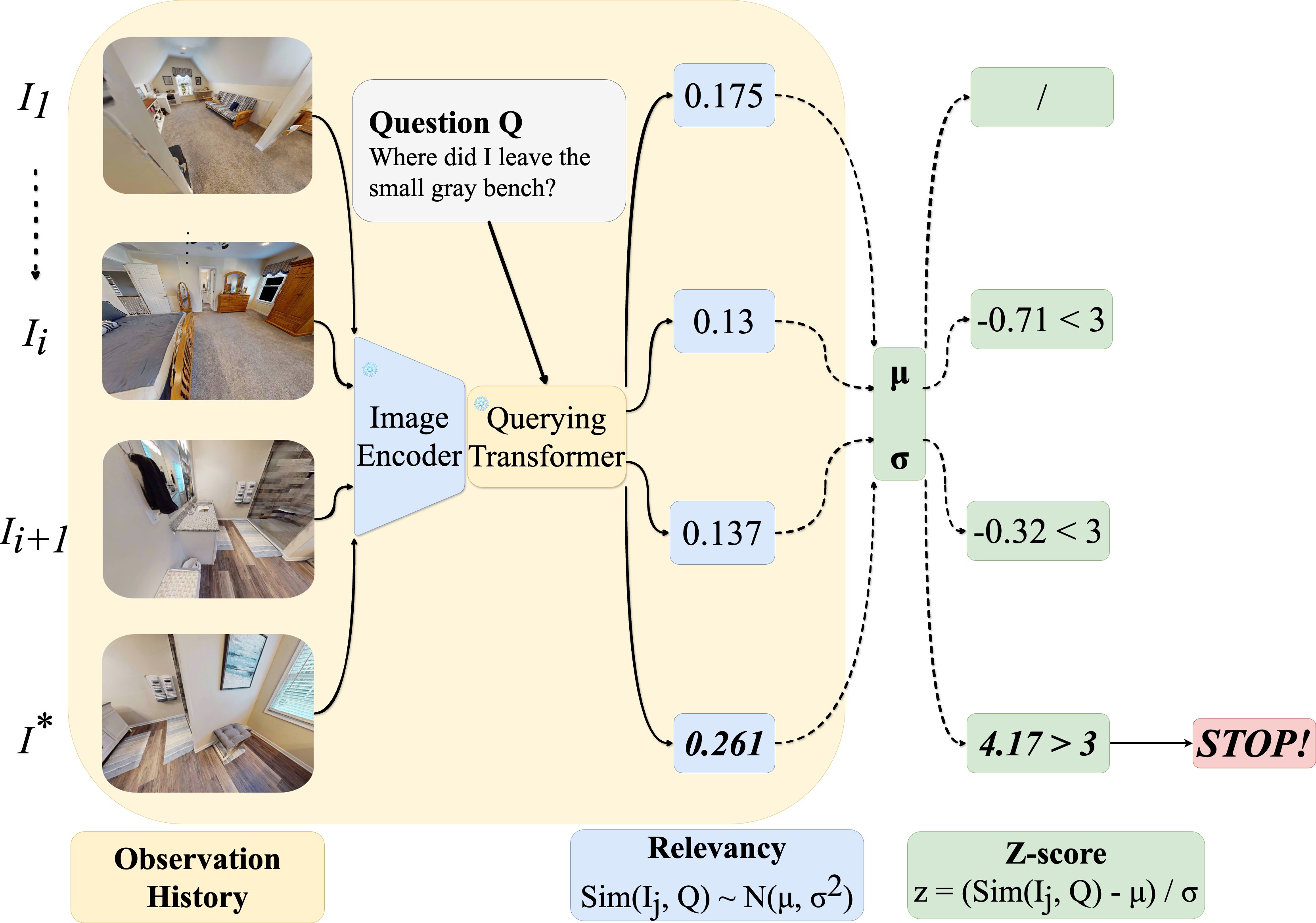}
    \caption{\textbf{Relevancy-Based Early Stopping Strategy}. In this example, we use $\widehat{\mu_I}+3\widehat{\sigma_I}$ as a threshold for detecting the \textit{answer-decisive} frame $H^*$ whose relevancy $\text{Sim}(I^*, Q)$ is an outlier for $\mathcal{N}(\mu_I, \sigma_I^2)$.}
    \label{fig:earlystop}
\end{figure}
We collect an episode history $\mathit{H} = \{H_1, H_2, ..., H_n\}$ for the $n$ steps' exploration of the robot agent, where each observation $H_t = (I_t, D_t)$ is the robot's observation taken at the $t$-th step's pose $P_t$, containing an RGB image $I_t \in \mathbb{R}^{H_I \times W_I \times 3}$ and a depth image $D_t \in \mathbb{R}^{H_I \times W_I}$. We compute the similarity score $\text{Sim}(I_t, Q)$ of the RGB image $I_t$ and the question $Q$ with BLIP~\cite{li2023blip} model.:
\begin{equation}
\label{eq:sim}
    \text{Sim}(I_t, Q) = \textbf{BLIP}(I_t, Q).
\end{equation}
We design an early stopping strategy for the agent's exploration by detecting an \textit{answer-decisive} frame $H^* \in H$, which is sufficiently related to the question $Q$ that the agent can provide a determined answer $A^*$ based on all observations collected up to the point $\{H_1, H_2, ...,H^*\}$, without requiring any further observations.

Assume the episode history $\mathit{H}' = \{H_1, H_2, ..., H_i\}$ up to the $i$-th step of the agent's exploration is not decisive for providing an answer. For $\{H_1, H_2, ..., H_i\}$, we assume their similarity scores with respect to $Q$ follow a normal distribution, expressed as:
\begin{equation}
    \text{Sim}(I_j, Q) \sim \mathcal{N}(\mu_I, \sigma_I^2) \quad \text{for} \quad j = 1, 2, \dots, i.
\end{equation}
In contrast, the \textit{answer-decisive} frame $H^*$ is so relevant to $Q$ that its similarity score $\text{Sim}(I^*, Q)$ is an outlier that is greater than $\mu_I+Z_{th}\sigma_I$ of this normal distribution. $Z_{th}$ is a hyperparameter to balance accuracy and efficiency. 

In practice, we calculate the sample mean \( \hat{\mu} \) and sample standard deviation \( \hat{\sigma} \)
  based on the available episode history $\mathit{H}'= \{H_1, H_2, ..., H_i\}$ as 
\begin{align}
    \widehat{\mu_I} =& \frac{1}{i} \sum_{j=1}^{i} \text{Sim}(I_j, Q),\\
    \widehat{\sigma_I} =& \sqrt{\frac{1}{i-1} \sum_{j=1}^{i} (\text{Sim}(I_j, Q) - \widehat{\mu_I})^2}.
\end{align}
We can then compute the probability that a new observation $H^*$ is an \textit{answer-decisive} frame. We use the Z-score $z$ to calculate the stopping threshold given by
\begin{equation}
    z = \frac{\text{Sim}(I_j, Q)- \widehat{\mu_I}}{\widehat{\sigma_I}}.
\end{equation}
Given $Z_{th}=3$, for a regular observation $H_j= (I_t, D_t)$ which satisfies $\text{Sim}(I_j, Q) \sim \mathcal{N}(\mu_I, \sigma_I^2)$, the probability that $H_j$ is misclassified as an \textit{answer-decisive} frame under the $\widehat{\mu_I}+3\widehat{\sigma_I}$ threshold is less than $1 - \Phi(3) = 0.13\%$, where \( \Phi(x) \) is the cumulative distribution function of a standard normal distribution. Thus, we can use $\widehat{\mu_I}+3\widehat{\sigma_I}$ as a threshold for detecting the \textit{answer-decisive} frame $H^*$ and stopping (cf. Fig.~\ref{fig:earlystop}). To avoid stopping due to early variances of few samples, our method will not stop before the 10th step.

\subsection{\textbf{Answering:} Retrieval-Augmented Generation for Question Answering}
\label{sec:ans}
\begin{figure}[t]
    \centering
    \includegraphics[width=0.8\linewidth]{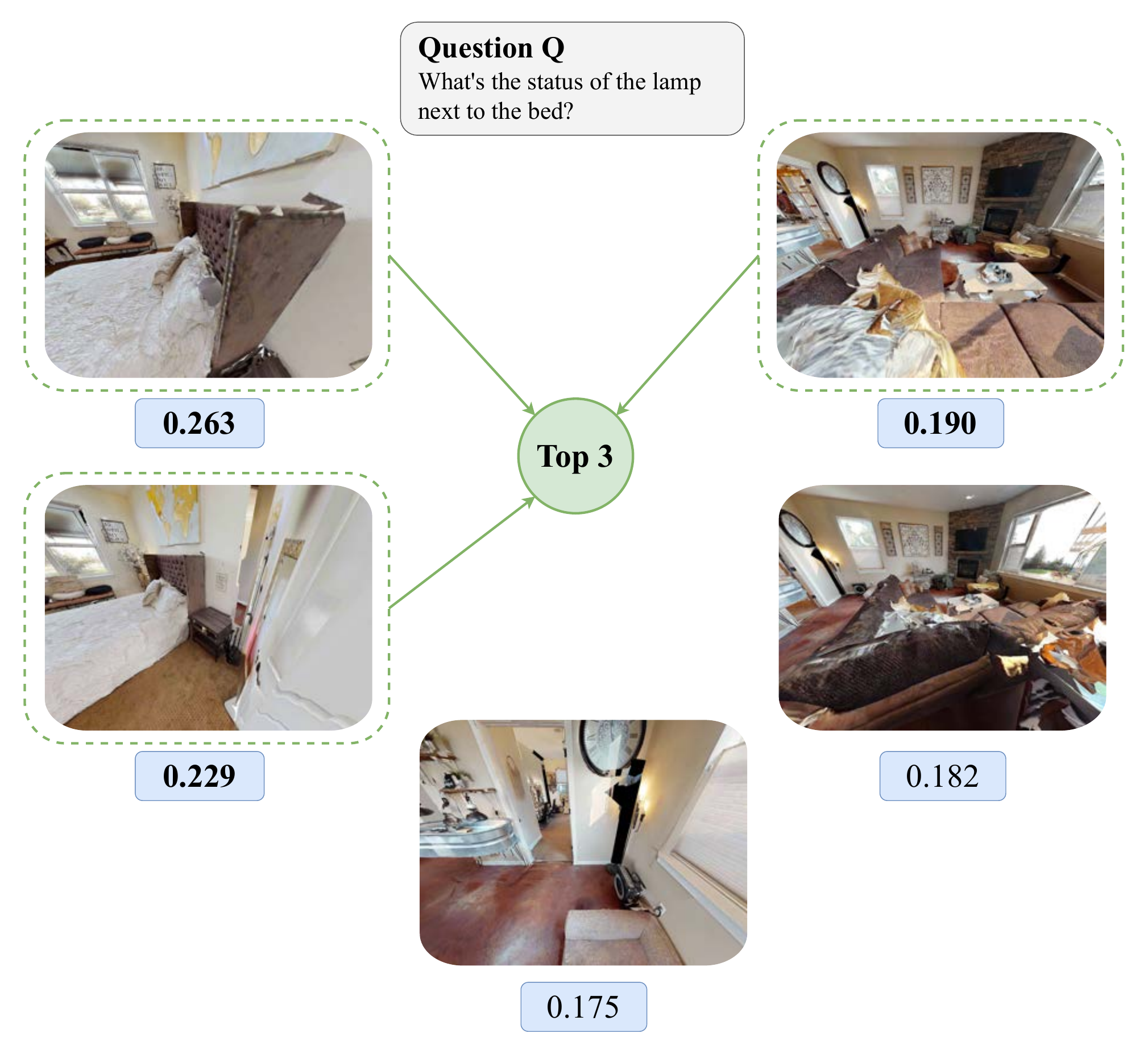}
    \caption{\textbf{BLIP-Based Retrieval.} Given the question $Q$: 
    \textit{"What's the status of the lamp next to the bed?"}, we use the image-question relevancy $\text{Sim}(I^t, Q)$ (as reflected in \textcolor{black}{blue} boxes) as the retrieval criterion for the observation history episode $\mathit{H} = \{H_1, H_2, ..., H_n\}$. The first 3 images, which score the highest, depict both the lamp and the bed, making them most relevant.}
    \label{fig:blip}
\end{figure}
After the robot halts, we use RAG to generate the final answer $\mathit{A}$ for $\mathit{Q}$. Previous works~\cite{exploreeqa2024,majumdar2024openeqa} either evenly sample observations during exploration or answer questions based on a single image. In order to save computation and minimize distractions, we draw on the idea of~\cite{lewis2020retrieval}, only feeding those images that can help answer the question to VLMs, \ie, the top $K$ images with the highest image-question relevancy, as is shown in Fig.~\ref{fig:blip}. The image-question relevancy $\text{Sim}(I^t, Q)$ is calculated by Equation~(\ref{eq:sim}). 
\section{Experiments}

\subsection{Settings}

\textbf{Simulation and datasets.} We use the Habitat-Sim physics simulator~\cite{puig2023habitat3} with the HM-EQA dataset~\cite{exploreeqa2024}, an embodied question-answering dataset featuring photo-realistic 3D scans from the Habitat-Matterport 3D Research Dataset (HM3D)~\cite{ramakrishnan2021habitat} along with relevant questions. We evaluate on the first
100-question subset (out of the total 500 questions) to keep the commercial VLM runs tractable. While the original dataset provides multiple answer choices for each question, we remove the choice sets in our open-vocabulary setting.

\textbf{Evaluation.} To evaluate the \textit{accuracy} of open-vocabulary EQA, we utilize the LLM-Match metric, which has demonstrated robustness in several prior studies~\cite{majumdar2024openeqa, liu2023mmbench}. Given a triplet consisting of a question, a reference answer, and the model's prediction, a large language model (LLM) is tasked with assigning a score $\sigma_i \in \{1,2,3,4,5\}$, where a higher score reflects a better match. The aggregated LLM-based correctness metric (LLM-Match) is defined as
\begin{equation}
\label{eq:correc}
\textit{LLM-Match Score} = \frac{1}{N}\sum_{i=1}^{N} \frac{\sigma_i -1}{4} \times 100\%   .
\end{equation}
The \textit{efficiency} is evaluated based on the \textbf{average steps} taken before the robot halts, a value that correlates with the number of VLM calls and the total runtime. 
As such, it serves as a highly suitable metric for assessing efficiency.

For the \textbf{choice of models}, we use the \textit{gpt-4o-2024-05-13}~\cite{openai_gpt4o} from OpenAI as our VLM backbone, with temperature $T$ set to 1. We choose the BLIP-2 model version \textit{Salesforce/blip2-itm-vit-g}~\cite{li2023blip} to gauge the relevance between images and questions.

For the \textbf{choice of hyperparameters}, the default value for the Z-score threshold $Z_{th}$ is 3 and for $K$ is 5.

\subsection{Baselines and Ablations}

While there are no direct baselines for comparison, we modified and adapted the text-only LLM and Explore-EQA~\cite{exploreeqa2024} to accomplish \textbf{Open Explore EQA} for evaluation:
\begin{itemize}
    \item \textbf{Blind LLM} that is provided only with the text question, without any visual input. This tests the inherent world knowledge encoded in the LLM's parameters. For fair comparison, we use OpenAI's \textit{gpt-4o-2024-05-13} model for this \Kai{pure QA} baseline~\cite{achiam2023gpt}.
    \item \textbf{ExploreEQA}~\cite{exploreeqa2024} that employs Semantic-Value-Weighted Frontier Exploration (SFE) but requires access to the logits from VLMs. It uses Prismatic VLMs~\cite{karamcheti2024prismatic} in a white-box setting. Notably, comparing this baseline with our method is not entirely fair, as our approach assumes a black-box setting for VLM usage, while ExploreEQA has access to the model’s internal logits.
    \item \textbf{Black-Box ExploreEQA}~\cite{exploreeqa2024} that uses verbalized confidence \Kai{scores} instead of logits to guide exploration, while all other components of ExploreEQA remain unchanged.
\end{itemize}

In addition, we {conduct an ablation study by comparing our method with two ablation variants, each of which evaluates the effect of changing a core component:}

\begin{itemize}
    \item 
\textbf{Ours with different stopping criteria} that uses different Z-scores as the stopping criteria for detecting the \textit{answer-decisive} frame.
\item \textbf{Ours with different RAG hyperparameters} that uses different RAG hyperparameters to filter the observation history episode $\mathit{H} = \{H_1, H_2, ..., H_n\}$.
\end{itemize}

\subsection{Results and Analysis}
\subsubsection{Comparisons with Baselines}

\begin{table}[htbp] 
  \centering
  \caption{\textbf{Quantitative Evaluations and Comparisons with Baselines.} In each entry, we report \textbf{Average Steps} and \textbf{LLM-Match Score(\%)} for \Kai{our task of Open Explore EQA}.}
    \begin{tabular}{@{}lcc@{}}
    \toprule
    Method              & Average Steps $\downarrow$ & LLM-Match ($\uparrow$, \%) \\ \midrule \midrule
    Blind LLM~\cite{achiam2023gpt} & - & 29.50 \\ \midrule
    ExploreEQA~\cite{exploreeqa2024}         & 38.87 &   {39.00}          \\ \midrule
    $\text{ExploreEQA}_{\text{Black Box}}$~\cite{exploreeqa2024}        & 38.87 &    36.50          \\ \midrule
    EfficientEQA (Ours)      & \textbf{30.16}  &  \textbf{54.25}    \\  \bottomrule
    \end{tabular}
  \label{tab:vs_baseline}
\end{table}

TABLE~\ref{tab:vs_baseline} shows that our method surpasses both baselines in terms of accuracy (LLM-Match) and efficiency (Average Steps). This improvement is partly due to the baselines' reliance on logits, which restricts their ability to leverage more powerful models like GPT-4o~\cite{achiam2023gpt}, whose logits are not always accessible. In contrast, our method seamlessly integrates black-box models and allows the VLM to provide comprehensive answers based on multiple relevant images. Notably, the blind LLM achieves decent performance, verifying that reasonable guesses can be made merely based on prior world knowledge.

In terms of efficiency, the baseline’s dependence on multi-step conformal prediction~\cite{ren2023robots} is a limiting factor, as it only applies to questions with predefined answer options and cannot terminate earlier than a set number of steps. Our method, instead, employs an early stopping mechanism, enabling it to stop once the correct answer is determined, \Kai{thus} enhancing efficiency.

\subsubsection{Comparisons with Ablations}

TABLE~\ref{tab:ab_1} shows our method's performance under different stopping thresholds $Z_{th}$. Our method always beats the baselines in TABLE~\ref{tab:vs_baseline} in both Average Steps and LLM-Match scores, which proves the robustness of our method. Intuitively, with an increasing $Z_{th}$, the framework should be stricter about when to stop, and thus expects a higher number in both Average Steps and LLM-Match scores. The exception in TABLE~\ref{tab:ab_1} that the Average Steps of $Z_{th}=2$ is greater than that of $Z_{th}=1$ should come from the variance in the exploration process with VLMs. For any scene unexplored, the robot agent will always need a minimum number of steps to reach the related target. The Average Steps for Ours ($Z_{th}=1$) and Ours ($Z_{th}=2$) are actually reaching that lower bound.

\begin{table}[bthp] 
  \centering
  \caption{\textbf{Ablation Study of Our Method with Different Stopping Criteria.} We tested our method using 3 different Z-score thresholds $Z_{th}$ as the stopping criteria \Kai{for detecting the \textit{answer-decisive} frame}.}
    \begin{tabular}{@{}lcc@{}}
    \toprule
    Method              & Average Steps $\downarrow$ & LLM-Match ($\uparrow$, \%)  \\ \midrule \midrule
    Ours ($Z_{th}=1$)        & 28.94 &   47.75        \\ \midrule
    Ours ($Z_{th}=2$)        & \textbf{28.70} &   53.50          \\ \midrule
    Ours ($Z_{th}=3$)      & 30.16  &  \textbf{54.25}    \\  \bottomrule
    \end{tabular}
  \label{tab:ab_1}
  \vspace{-6mm}
\end{table}

\begin{table}[h!]
  \caption{\textbf{Ablation Study of Our Method with Different RAG Hyperparameters.} $K$ represents the number of observations fed into the VLM for RAG. We tested our method using 4 different hyperparameters $K$.
  }
  \centering
    \begin{tabular}{@{}lc@{}}
    \toprule
    Method              &  LLM-Match ($\uparrow$, \%) \\ \midrule \midrule
    Ours ($K=10$)         &   53.50        \\ \midrule
    Ours ($K=5$)         &   \textbf{54.25}          \\ \midrule
     Ours ($K=3$)         &   52.00          \\ \midrule
    Ours ($K=1$)        &  51.25   \\  \bottomrule
    \end{tabular}
  \label{tab:ab_2}
\end{table}

\begin{figure}[tbp]
    \centering
    \includegraphics[width=0.95\linewidth]{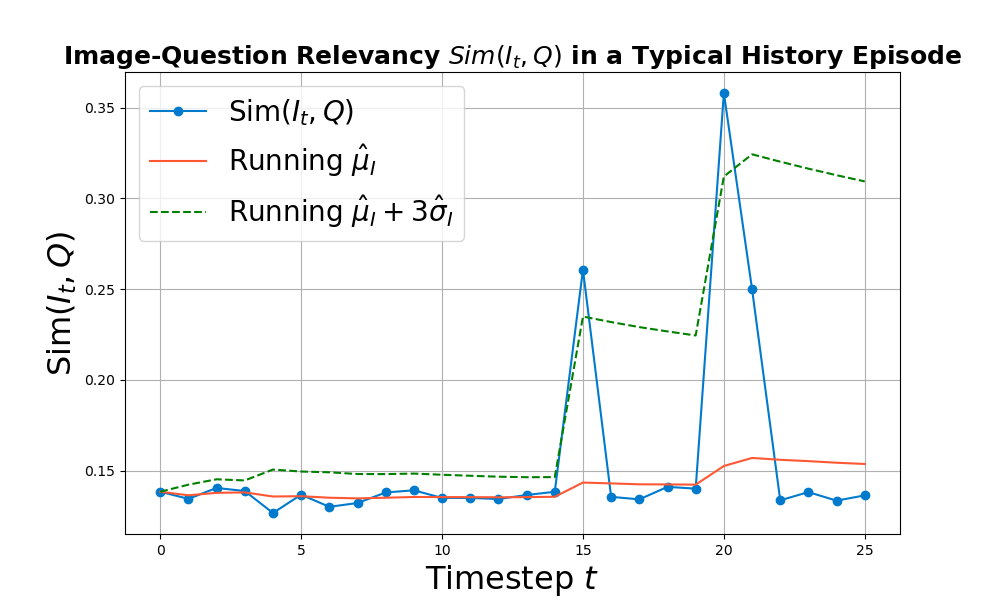}
    \caption{\textbf{Distribution of the Image-Question Relevancy $\textit{Sim}(I_t, Q)$ in a Typical History Episode}. The episode is collected without early stopping for illustration. Most samples fall under the threshold of $\widehat{\mu_I}+3\widehat{\sigma_I}$ except for the \textit{answer-decisive} frame $H^*$. The observations of the two peak values can be seen in Fig.~\ref{appfig:visu_peak}. With our early stopping strategy, the agent would be able to stop at the 15th step. 
    }
    \vspace{-2mm}
    \label{fig:rel}
\end{figure}

\begin{figure}[tbp]
    \centering
    \begin{minipage}[b]{0.23\textwidth}
        \centering
        \includegraphics[width=\linewidth]{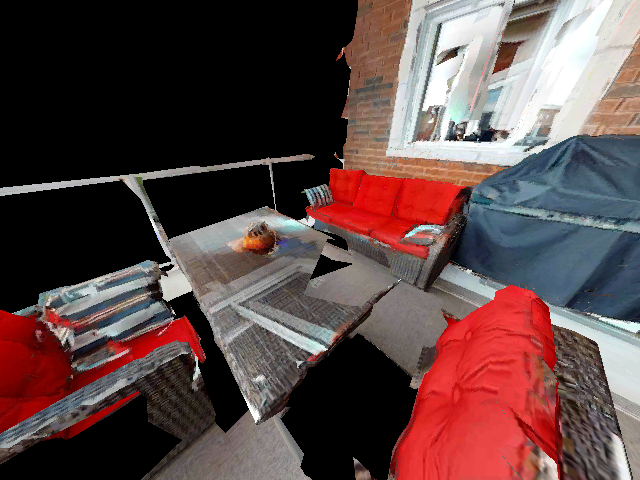}
    \end{minipage}
    \hfill
    \begin{minipage}[b]{0.23\textwidth}
        \centering
        \includegraphics[width=\linewidth]{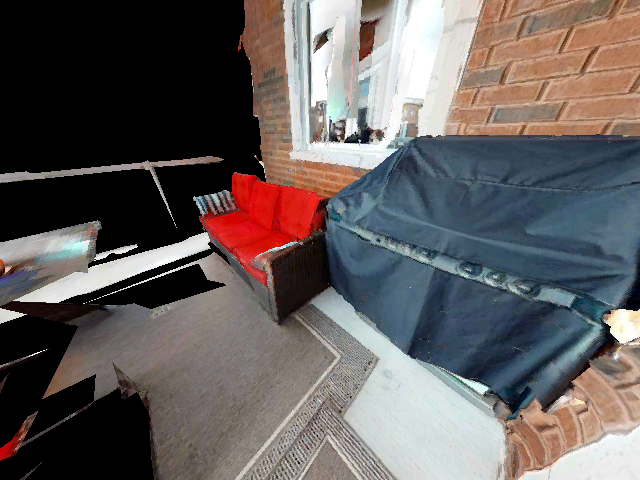}
    \end{minipage}
    \caption{\textbf{Observations of the Two Image-Question Relevancy Peaks in Fig.~\ref{fig:rel}}. These two images are the 15th and 20th steps from the history episode respectively. The question $Q$ is 
    \textit{"What is on the outdoor grill?"}
    }
    \vspace{-7mm}
    \label{appfig:visu_peak}
\end{figure}

Fig.~\ref{fig:rel} visualizes the distribution of image-question relevancy $\textit{Sim}(I_t, Q)$ across a typical historical episode. During exploration, the majority of observations do not significantly contribute to answering the question, which helps in establishing the running statistics. The retrieval performance of the BLIP model is robust, consistently yielding high scores for images closely related to the question. This reliability supports the viability of our proposed early stopping methodology. The observations of the two peak image-question relevancy from Fig.~\ref{fig:rel} are shown in Fig.~\ref{appfig:visu_peak}.

TABLE~\ref{tab:ab_2} presents the LLM-Match scores under varying numbers of considered observations within the VLM. Interestingly, using only the most pertinent image ($K=1$) achieves a respectable level of correctness. The optimal performance is observed at $K=5$, suggesting a balance between relevant information and cognitive overload. However, as the number increases to $K=10$, there is a noticeable decline in performance, likely due to the inclusion of less relevant or distracting images that may mislead VLM's decision-making.

\section{Conclusions}

We formulate the novel task of \textbf{Open Explore EQA} \Kai{(open vocabulary embodied question answering during exploration) with motivation from real-world scenarios}. Using Semantic-Value-Weighted Frontier Exploration 
\Kai{together with black-box VLMs}, we develop a novel framework, \textbf{EfficientEQA}, for open vocabulary embodied question answering. We further propose the Retrieval-Augmented Generation-based VLM querying method and outlier detection-based early stopping strategy to improve our method's accuracy and efficiency. Comprehensive experiments using the Habitat-Sim physics simulator~\cite{puig2023habitat3} equipped with large-scale Habitat-Matterport 3D Research Dataset (HM3D)~\cite{ramakrishnan2021habitat} and HM-EQA~\cite{exploreeqa2024} validated the effectiveness of our proposed approach.


\bibliographystyle{plain}
\bibliography{reference.bib}

\end{document}